\documentclass[10pt,journal,compsoc]{IEEEtran}

\usepackage{amsmath,amssymb}
\usepackage{graphicx}
\usepackage{booktabs}
\usepackage{algorithm}
\usepackage{algpseudocode}
\usepackage{xcolor}
\usepackage{url}
\usepackage{tikz}
\usetikzlibrary{positioning,arrows.meta}
\usepackage[hidelinks]{hyperref}

\ifCLASSOPTIONcompsoc
  \usepackage[nocompress,sort]{cite}
\else
  \usepackage{cite}
\fi

\providecommand{\tightlist}{\setlength{\itemsep}{0pt}\setlength{\parskip}{0pt}}

\begin{document}
\title{Faithful Action-unit Causal Reasoning for Counterfactually Faithful Emotion Explanations}

\author{Van Thong~Huynh, Hong Hai~Nguyen,~
        Thuy~Pham,
        Trong Nghia~Nguyen,~
        and~Soo-Hyung~Kim
\IEEEcompsocitemizethanks{\IEEEcompsocthanksitem Van Thong Huynh and Thuy Pham are with Faculty of CSE, Ho Chi Minh City University of Technology (HCMUT), VNUHCM, Vietnam. Hong Hai Nguyen is with Dept. of AI, FPT University, Vietnam. Trong Nghia Nguyen is with Faculty of DSAI, College of Technology, National Economic University, Vietnam. Soo-Hyung Kim is with Dept. of AI Convergence, Chonnam National University,
South Korea.\protect\\
\IEEEcompsocthanksitem Corresponding author: Van Thong~Huynh.}

}

\markboth{}
{FACR: Faithful Action-unit Causal Reasoning}

\IEEEtitleabstractindextext{%
\begin{abstract}
Multimodal models can name the action units (AUs) behind a facial
emotion, but their AU$\to$emotion rationales are typically plausible
rather than faithful: nothing forces the AUs a model invokes to
be the AUs that actually drive its prediction. We cast AU$\to$emotion
reasoning as a counterfactual-consistency problem between the rationale,
the label, and a structural AU$\to$emotion causal graph \(G\), and propose
FACR, which grounds the reasoner in an independently induced,
polarity-aware \(G\) and trains a counterfactual-faithfulness objective:
a do-intervention on an AU that \(G\) marks causal for a class must move
the prediction, while one it marks irrelevant must leave it unchanged.
Faithfulness is thereby both trainable and measurable through a matching
interventional metric, which we evaluate against a known causal
structure, the PSPI pain-AU composition, as no existing
affective-reasoning benchmark allows. We are explicit that this metric
tests fidelity to the supplied structure rather than its rediscovery: it
asks whether the trained reasoner invokes the AUs the structure marks
causal, on held-out subjects and a second dataset. Under
subject-independent (leave-one-subject-out) evaluation on UNBC-PAIN, the
objective raises the agreement between the invoked AUs and the PSPI
composition from a no-objective baseline of 0.08 to 0.57, at a small
detection cost (AUC 0.83 to 0.79); an unfaithfulness control attributes
the gain to the objective. On a cross-dataset emotion transfer
(AffectNet $\times$ FABA $\to$ Aff-Wild2), the objective likewise raises fidelity
to \(G\) on a seven-class task (0.50 to 0.84). Agreement with the
truly active AUs is a separate, stricter test, and it is bounded
by the graph\textquotesingle s correctness: with the learned graph it
stays at the control level (0.24), and only a verified-graph (EMFACS)
variant, corroborated by a corrupted-graph control, raises it to
0.59. Finally, we attach a language verbalizer and extend the audit to
the generated text: biasing each action unit\textquotesingle s emission
by its latent activation makes the rationale faithful by construction,
so that ablating an AU removes it from the explanation, a property
that transfers to a second language-model backbone, whereas a freely
generated rationale is unfaithful.

\end{abstract}

}

\maketitle

\IEEEdisplaynontitleabstractindextext
\IEEEpeerreviewmaketitle

\ifCLASSOPTIONcompsoc
\IEEEraisesectionheading{\section{Introduction}\label{sec:introduction}}
\else
\section{Introduction}
\label{sec:introduction}
\fi

Automatic recognition of facial emotion is a central problem in
affective computing, with applications in human--computer interaction,
driver monitoring, and health. As these systems move into
decision-sensitive settings, a predicted label is no longer sufficient
on its own; the system is increasingly expected to explain why an
emotion was inferred, in terms a person can check. For facial behavior,
the natural vocabulary for such an explanation is the facial action unit
(AU), the elementary muscle movement of the Facial Action Coding System,
since specific AU combinations are the physical evidence of an
expression. An explanation is useful, however, only when it is faithful, that is, when it reflects the factors that actually drive the
prediction, rather than a plausible account that the model did not rely
on.

Two families of methods connect AUs to emotion, but neither delivers a
faithful explanation. Structural-causal methods model AUs and
expressions with explicit causal graphs and interventions, and have been
used to deconfound subjects \cite{ref1}, debias context \cite{ref2}, and
discover directed AU dependencies \cite{ref3,ref6}; nevertheless, they are
purely discriminative and expose only a label or an attention map,
discarding the recovered structure at inference. Comparatively,
multimodal large language models (MLLMs) have been adapted to verbalize
AU-based emotion rationales \cite{ref7,ref8,ref9,ref13}; however, the generated
text is not constrained by any verified causal structure and is
evaluated only for plausibility, so it is often unfaithful,
substituting causal attribution with template matching and failing to
update under counterfactual edits \cite{ref15,ref16}. Since the
explanation and the prediction are produced by decoupled mechanisms,
intervening on an AU that is decisive for a class does not reliably
change both the rationale and the label together, which is precisely the
behavior a trustworthy reasoner should exhibit.

In this study, we focus on grounding an MLLM emotion reasoner in a
structural AU$\to$emotion causal graph, so that its explanation is
counterfactually faithful rather than merely plausible. The proposed
system, Faithful Action-unit Causal Reasoning (FACR), can be split into
three stages: 1) we induce a directed, polarity-aware AU$\to$emotion causal
graph \(G\) by independent structure discovery, anchored where ground
truth exists by the pain-intensity AU composition \cite{ref34}; 2) we
condition the reasoner on \(G\), so that the prediction is routed
through a disentangled AU latent and the rationale is generated over the
graph\textquotesingle s nodes and edges; and 3) we introduce a
counterfactual-faithfulness objective that applies do-interventions in
the AU latent space, requiring the prediction and rationale to change
for AUs that \(G\) marks causal for a class and to stay invariant for
AUs it marks irrelevant. Based on that, the faithfulness of the
explanation becomes both trainable, through the objective, and
measurable, through a matching interventional metric that we evaluate
against the known pain AU composition on UNBC-PAIN. The metric tests
whether the trained reasoner is faithful to the supplied structure on
held-out data, not whether it rediscovers that structure from scratch;
this distinction is what makes the score interpretable, and we return to
it in \autoref{sec:disc}.

We emphasize that the individual ingredients each appear in prior
work \cite{ref6,ref13,ref14,ref32}: an AU-grounded reasoner, a causal AU
graph, and a counterfactual constraint. The contribution of this study
is their specific combination in the AU$\to$emotion setting and, in
particular, a faithfulness criterion that can be evaluated against a
known causal structure, which no existing affective-reasoning benchmark
provides. Experimental results on UNBC-PAIN and a cross-dataset emotion
transfer to Aff-Wild2 show that the faithfulness objective substantially
increases the agreement between the explanation and the causal graph,
and, when the graph is verified, the agreement with the truly active AUs, at a small, quantified cost to detection accuracy.

In summary, the main contributions of this study are as follows:

\begin{itemize}
\tightlist
\item
  We formalize the faithfulness gap between structural AU causal models
  and MLLM emotion reasoning, and cast AU$\to$emotion reasoning as a
  counterfactual-consistency problem between the rationale, the label,
  and a causal graph.
\item
  We propose FACR, which grounds the reasoner in an independently
  induced, polarity-aware AU$\to$emotion causal graph and enforces
  faithfulness through do-interventions in a disentangled AU latent
  space.
\item
  We introduce a counterfactual faithfulness metric for verbalized
  AU$\to$emotion explanations and evaluate the invoked AUs against the PSPI
  pain composition on UNBC-PAIN, as a test of fidelity to a known
  structure rather than its rediscovery.
\item
  We demonstrate that the constraint improves faithfulness to \(G\) on
  the UNBC-PAIN anchor and a cross-dataset emotion transfer, attribute
  the gain to the objective with an unfaithfulness control, and show
  that agreement with the truly active AUs is a separate test
  bounded by the graph\textquotesingle s correctness, improving only
  when the graph is verified, at a modest, quantified cost to
  recognition accuracy.
\end{itemize}

\section{Related Work}\label{related-work}

Two largely separate lines of work connect facial muscle activity to
emotion: one recovers an explicit causal structure over action units
(AUs) but uses it only for discriminative prediction, and the other
emits natural-language emotion explanations but leaves them
unconstrained by any verified structure. We review each in turn, then
the faithfulness tools our metric builds on, and close with the gap that
motivates FACR.

\subsection{Structural causal models for AUs and
expressions}\label{structural-causal-models-for-aus-and-expressions}

A growing body of work introduces causal structure into AU and
facial-expression recognition to remove spurious correlations. In
\cite{ref1}, the authors formulated AU recognition as a structural causal
model and proposed a plug-in intervention module that deconfounds the
subject variable through backdoor adjustment. Yang et al. \cite{ref2}
modeled context-aware emotion recognition with a causal graph and
debiased the prediction by subtracting the direct context effect,
comparing factual and counterfactual outcomes at inference. Tan et al.
\cite{ref3} replaced AU co-occurrence adjacency with a discovered AU
causal graph for micro- and macro-expression spotting, which suppressed
dataset-biased AU--emotion links, while CoDER \cite{ref4} learned the
temporal-relationship effect of micro-expressions by contrasting factual
and counterfactually-revised sequences. Causal reasoning has likewise
been employed for modality debiasing in multimodal emotion recognition
\cite{ref5}. Most relevant to our work, CausalAffect \cite{ref6}
discovered a directed, polarity-aware causal graph over AU$\to$AU and
AU$\to$expression dependencies and enforced it through a feature-level
counterfactual intervention in a disentangled AU latent space.
Comparatively, all of these methods are purely discriminative: the
recovered structure improves a label or an attention map, and none
produces or audits a natural-language explanation.

\subsection{Multimodal large language models for emotion
reasoning}\label{multimodal-large-language-models-for-emotion-reasoning}

In parallel, multimodal large language models (MLLMs) have been adapted
to emit natural-language emotion rationales. FEALLM \cite{ref7}
constructed an instruction dataset that aligns facial expressions with
AU descriptions and verbalizes their relationship, and EmoLA \cite{ref8}
instruction-tuned an MLLM for joint emotion and AU recognition,
evaluating the generated text with a recognition-plus-generation metric.
Emotion-LLaMA \cite{ref9} and AffectGPT \cite{ref10} scaled multimodal
emotion reasoning with larger instruction corpora, while EMO-LLaMA
\cite{ref11} incorporated AUs at the feature level and XEmoGPT
\cite{ref12} grounded explanations in modality-specific cues. More recent
systems tie the reasoning more tightly to AUs: TAG \cite{ref13}
constrained the reasoning steps to AU-related facial regions through a
reinforcement reward, and AULLM++ \cite{ref14} combined an LLM reasoner
with a structural AU-graph prior and a counterfactual-consistency
regularizer for micro-expression AU detection. Nevertheless, these
systems evaluate their explanations by recognition accuracy or by
plausibility, text overlap, a GPT-based judge, or semantic cue
matching, rather than by whether the explanation reflects the factors
that actually drive the prediction.

\subsection{Faithfulness of
explanations}\label{faithfulness-of-explanations}

Whether an explanation is faithful, that is, whether it
reflects the model\textquotesingle s true reasoning rather than merely a
plausible account, is a distinct and well-studied question outside
affective computing. Jacovi and Goldberg \cite{ref16} formalized the
faithfulness-versus-plausibility distinction, and Turpin et al.
\cite{ref15} showed that chain-of-thought rationales can be plausible yet
unfaithful, shifting under biasing interventions that the explanation
never mentions; RFEval \cite{ref19} further reported that
reinforcement-style objectives can reduce reasoning faithfulness while
accuracy holds. A complementary line provides interventional
faithfulness criteria that our metric draws on: comprehensiveness and
sufficiency through rationale erasure \cite{ref20}, the
counterfactual-edit test of Atanasova et al. \cite{ref22}, counterfactual
simulatability \cite{ref21}, and counterfactual invariance to irrelevant
input changes \cite{ref17}. For high-level concepts specifically, the
causal concept effect \cite{ref25} measured the do-intervention effect of
a concept on a prediction, in contrast to the correlational sensitivity
of concept activation vectors \cite{ref26}, and causal mediation analysis
\cite{ref28} distinguished information that merely exists in a
representation from information the model actually uses. These ideas
were recently extended to vision-language explanations: EDCT \cite{ref24}
audited a vision-language model\textquotesingle s natural-language
explanation by counterfactual image edits, and CC-SHAP \cite{ref23}
measured self-consistency between answer and explanation attributions.
Two further connections are worth noting. Predicting a label from
intermediate concepts and intervening on those concepts is the
concept-bottleneck paradigm \cite{ref29,ref30}; FACR differs in that it
audits a verbalized rationale rather than a hard concept vector.
Regularizing a model\textquotesingle s reasoning to be causally
consistent has likewise been studied for large-language-model reasoning
\cite{ref32} and for selective rationales \cite{ref33}, and FACR
specializes this idea to AU-grounded emotion reasoning, defining both a
sensitivity and an invariance arm from a verified causal graph and
evaluating it against a known structure.

\subsection{The gap}\label{the-gap}

Most of the previous studies fall on one side of a divide.
Structural-causal methods \cite{ref1,ref2,ref3,ref4,ref6} recover a graph but discard
it at inference and never verbalize a rationale, whereas affective MLLMs
\cite{ref7,ref8,ref9,ref10,ref11,ref12,ref13} verbalize a rationale yet leave it
unconstrained by a verified structure and evaluate it only for
plausibility, text overlap or a model-based judge. The verbalized
AU$\to$emotion reasoning is therefore often unfaithful, substituting causal
attribution with template matching and failing to update under
counterfactual edits \cite{ref15,ref16}. Closing this gap is non-trivial for
two reasons. First, a trustworthy AU$\to$emotion structure is needed,
because co-occurrence mappings carry dataset and demographic bias
\cite{ref3}; we therefore induce the graph and anchor it where a
ground-truth composition exists. Second, the rationale must be tied to
that structure through an interventional signal that makes faithfulness
both trainable and measurable \cite{ref20,ref22}, rather than scored by
surface similarity.

The works closest to ours each share one ingredient but not the
combination. CausalAffect \cite{ref6} discovers the causal AU graph and
intervenes in an AU latent space, but is discriminative and audits no
explanation. AULLM++ \cite{ref14} pairs an LLM with an AU-graph prior and
a counterfactual-consistency regularizer, but the prior is learned
rather than verified, the regularizer targets generalization rather than
a faithfulness audit, and it generates no rationale. EDCT \cite{ref24}
audits a vision-language explanation by counterfactual edits, but in
general visual question answering, without a causal graph and as an
evaluation only. Concept-bottleneck models \cite{ref29,ref30} predict from
intervenable concepts, but expose a concept vector rather than a
verbalized rationale and judge intervention by accuracy gain, not by a
sensitivity--invariance signature. To our knowledge, no prior work
grounds a natural-language AU$\to$emotion reasoner in a verified
causal graph and audits the counterfactual faithfulness of its
explanation against a known structure; this intersection is where FACR
sits. We state this as positioning, verified against the recent
literature, rather than as a claim of priority for any single mechanism.

\section{Method}\label{method}
\begin{figure*}[t]
\centering
\begin{tikzpicture}[
  node distance = 9mm and 13mm,
  box/.style = {draw, rounded corners, minimum height=9mm, minimum width=15mm, align=center,
                fill=gray!8, font=\small},
  io/.style  = {align=center, font=\small},
  >=Latex]
  \node[io]  (img)  {Face\\image};
  \node[box, right=of img] (enc) {ResNet-18\\encoder};
  \node[box, right=of enc] (z)   {AU latent\\$z$};
  \node[box, right=of z, yshift=7.5mm]  (head) {graph-cond.\\emotion head};
  \node[box, right=of z, yshift=-7.5mm] (lm)   {$z$-gated LM\\verbalizer};
  \node[io, right=of head] (emo) {emotion $\hat{y}$};
  \node[io, right=of lm]   (rat) {rationale\\(named AUs)};
  \node[box, below=8mm of z, fill=blue!6] (G) {causal graph $G$};
  \node[font=\footnotesize, above=6mm of z] (do) {$\mathrm{do}(z_k{=}0)$};

  \draw[->] (img)--(enc);  \draw[->] (enc)--(z);
  \draw[->] (z.east) -- (head.west);  \draw[->] (z.east) -- (lm.west);
  \draw[->] (head)--(emo);  \draw[->] (lm)--(rat);
  \draw[->, dashed] (do)--(z);
  \draw[->, dotted] (G) to[bend left=18]  (head.south);
  \draw[->, dotted] (G) to[bend right=18] (lm.south);
\end{tikzpicture}
\caption{FACR overview. A face image is encoded into a disentangled action-unit (AU) latent
$z$; a graph-conditioned head predicts the emotion and a $z$-gated language model verbalizes
the rationale over the AUs. The causal graph $G$ defines, per class, which AUs are causal.
The counterfactual-faithfulness objective requires a do-intervention $\mathrm{do}(z_k{=}0)$
on a graph-causal AU to move both the prediction and the rationale, and on an irrelevant AU
to leave them unchanged; gating each AU's emission on $z$ makes the verbalized rationale
faithful by construction.}
\label{fig:arch}
\end{figure*}
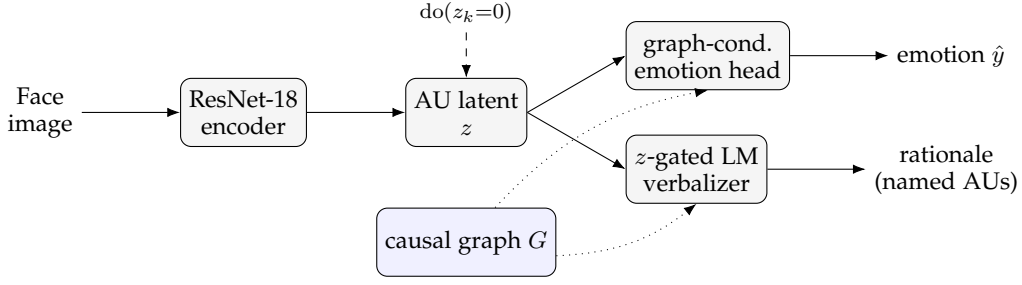

This section describes the key components of our system in detail. The
proposed system, Faithful Action-unit Causal Reasoning (FACR), can be
split into three main parts: a structural AU$\to$emotion causal graph, a
graph-conditioned reasoner, and a counterfactual-faithfulness objective
that couples the reasoner\textquotesingle s rationale and prediction to
the graph, as illustrated in \autoref{fig:arch}.

\subsection{Overview and notation}\label{overview-and-notation}

\autoref{fig:arch} gives an overview. Let an input face be \(x\), with a set of
AU nodes \(A = \{a_1, \ldots, a_K\}\) and a set of emotion (and pain)
classes \(C\). The structural graph \(G = (A \cup C, E)\) is a directed,
polarity-aware graph whose edges \(E\) are either AU$\to$AU or AU$\to$class;
each edge carries a polarity \(s \in \{-1, +1\}\) (excitatory or
inhibitory) and a weight. For a class \(c\), we write \(\mathrm{Pa}(c)\)
for its direct causal AUs in \(G\). The reasoner maps \(x\) to a
disentangled AU latent \(z \in \mathbb{R}^K\), one coordinate per AU
node, and to a class prediction; \(z\) is the structured rationale over
which faithfulness is defined.

\subsection{Structural AU\texorpdfstring{$\to$}{->}emotion causal graph
induction}\label{structural-auemotion-causal-graph-induction}

We obtain \(G\) by independent discovery rather than by reusing an
external graph, so the method is self-contained and reproducible. The
graph is assembled from three sources, each tagged with its provenance.
AU$\to$emotion edges are estimated by an \(L_1\)-regularized logistic
regression of the class label on the AU activations, where a non-zero
coefficient yields a directed AU$\to$class edge whose sign defines its
polarity; this is fitted on the joint AU--emotion corpus formed by the
AffectNet $\times$ FABA join, and we note that its AU annotations are
model-generated, which we treat as a limitation. AU$\to$AU edges are
estimated by neighborhood selection, an \(L_1\)-logistic regression
of each AU on the remaining AUs, on EmotioNet \cite{ref37}, whose AU
labels are automatic. The pain sub-structure is fixed from the
Prkachin--Solomon pain-intensity (PSPI) formula \cite{ref34},
\begin{equation*}
    \mathrm{PSPI} = \mathrm{AU4} + \max(\mathrm{AU6},\mathrm{AU7}) + \max(\mathrm{AU9},\mathrm{AU10}) + \mathrm{AU43},
\end{equation*}
which provides a verified partial ground truth. Edges that the discovery
does not cover are filled from the EMFACS prototypes. As a
sanity check on the induced graph rather than as a result, we note that
the discovered AU$\to$emotion edges recover EMFACS-consistent structure (for
example, happiness $\leftarrow$ AU6, AU12; disgust $\leftarrow$ AU9, AU10).
\begin{algorithm}[t]
\caption{FACR training (staged)}\label{alg:facr}
\begin{algorithmic}[1]
\Require images $x$ with AU labels $a$ and class labels $y$; graph $G$ with per-class causal
sets $\mathrm{Pa}(c)$ and edge polarities $s_{k,c}\in\{-1,+1\}$; weights
$\lambda_a,\lambda_s,\lambda_i$; margin $\gamma$; gate scale $\beta$
\Statex \textbf{Stage 1 --- graph-conditioned reasoner (encoder $f_{\mathrm{enc}}$ + head)}
\For{minibatch $(x,a,y)$}
  \State $z \gets \sigma(f_{\mathrm{enc}}(x))$ \Comment{AU latent, $z_k\in(0,1)$}
  \State $\hat{y} \gets \mathrm{head}(z)$;\quad
         $\mathcal{L} \gets \mathcal{L}_{\mathrm{cls}}(\hat{y},y) + \lambda_a\,\mathcal{L}_{\mathrm{au}}(z,a)$
  \For{class $c$, AU $a_k$} \Comment{do-interventions in the AU latent}
     \State $\hat{y}^{\,\mathrm{do}(z_k=0)} \gets \mathrm{head}(z \text{ with } z_k{:=}0)$
     \If{$a_k \in \mathrm{Pa}(c)$}
        \State $\mathcal{L} \mathrel{+}= \lambda_s\,\big[\gamma - s_{k,c}\,(\hat{y}_c-\hat{y}_c^{\,\mathrm{do}(z_k=0)})\big]_+$ \Comment{sensitivity}
     \Else
        \State $\mathcal{L} \mathrel{+}= \lambda_i\,(\hat{y}_c-\hat{y}_c^{\,\mathrm{do}(z_k=0)})^2$ \Comment{invariance}
     \EndIf
  \EndFor
  \State update $f_{\mathrm{enc}},\mathrm{head}$ by $\nabla\mathcal{L}$
\EndFor
\State \textbf{freeze} $f_{\mathrm{enc}},\mathrm{head}$
\Statex \textbf{Stage 2 --- $z$-gated verbalizer (soft prompt + AU-token embeddings)}
\For{minibatch $(x,a,y)$}
  \State $z \gets \sigma(f_{\mathrm{enc}}(x))$ \Comment{no gradient to the encoder}
  \State bias each AU token's logit by $\beta\,\min(z_k-\tfrac12,\,0)$ \Comment{suppress inactive AUs}
  \State generate the rationale; set its stated emotion to $\arg\max_c \hat{y}_c$
  \State update the soft prompt and AU-token embeddings by $\nabla\mathcal{L}_{\mathrm{lm}}$
\EndFor
\end{algorithmic}
\end{algorithm}

\subsection{Graph-conditioned
reasoning}\label{graph-conditioned-reasoning}

The reasoner is designed so that both the prediction and the
natural-language rationale are routed through \(G\). An encoder maps the
input to a disentangled AU latent,

\begin{equation}
z = \sigma(f_{\mathrm{enc}}(x)),
\end{equation}
where \(\sigma\) is the sigmoid and each coordinate \(z_k \in (0, 1)\)
is the presence of AU \(a_k\); this latent is the interpretable
interface at which interventions act, since removing AU \(a_k\) is the
well-defined operation \(\mathrm{do}(z_k = 0)\). Conditioned on \(z\)
and on the graph, the reasoner produces two coupled outputs: a class
prediction \(\hat{y}\), and a rationale \(r\) that names the active AUs,
follows their AU$\to$class edges in \(G\), with the edge polarity stating
whether each AU supports or opposes the class, and concludes with the
predicted emotion. Specifically, the reasoner is conditioned on the
active AUs and their incident graph edges, so the rationale is generated
over the structure rather than alongside it; faithfulness is then
defined on this rationale together with the label (\autoref{sec:obj}). In the
minimal instantiation used for the latent-level audit (Sections \ref{sec:data}--\ref{sec:xdata}), the natural-language reasoner is replaced by a lightweight
class head on the same AU latent \(z\), and the faithfulness objective
is applied unchanged; because the objective depends only on \(z\) and
\(\hat{y}\), it transfers directly to the full reasoner.

\subsection{Counterfactual-faithfulness
objective}\label{sec:obj}\label{counterfactual-faithfulness-objective}

The central idea is that, under a do-intervention on an AU, the
prediction and the rationale must change together and consistently with
\(G\). We define two terms. The sensitivity term requires that
removing an AU that \(G\) marks causal for the target class \(c\) move
the class logit in the polarity-implied direction,
\begin{equation}
\mathcal{L}_{\mathrm{sens}} = \operatorname*{mean}_{a_k \in \mathrm{Pa}(c)} \big[\, \gamma - s_{k,c}\,\big(\hat{y}_c - \hat{y}_c^{\,\mathrm{do}(z_k=0)}\big) \,\big]_+ ,\end{equation}
where \(s_{k,c}\) is the edge polarity, \(\gamma\) the margin, and
\([\cdot]_+\) the hinge. The invariance term requires that
removing an AU that is irrelevant to \(c\) leave the class logit
unchanged,

\begin{equation}\mathcal{L}_{\mathrm{inv}} = \operatorname*{mean}_{a_k \notin \mathrm{Pa}(c)} \big(\hat{y}_c - \hat{y}_c^{\,\mathrm{do}(z_k=0)}\big)^2 .\end{equation}
The full objective adds these to the recognition loss and an
AU-grounding loss,

\begin{equation}\mathcal{L} = \mathcal{L}_{\mathrm{cls}} + \lambda_a \mathcal{L}_{\mathrm{au}} + \lambda_s \mathcal{L}_{\mathrm{sens}} + \lambda_i \mathcal{L}_{\mathrm{inv}} ,\end{equation}
where \(\mathcal{L}_{\mathrm{cls}}\) is the cross-entropy on the class,
\(\mathcal{L}_{\mathrm{au}}\) is the binary cross-entropy that grounds
the AU latent \(z\) in the AU labels, and
\(\lambda_a, \lambda_s, \lambda_i\) weight the terms. We use the
direct causal set \(\mathrm{Pa}(c)\) rather than the transitive
closure, so that the learned, and necessarily noisy, AU$\to$AU edges do not
contaminate a class\textquotesingle s causal set; in particular the pain
class retains exactly the PSPI AUs. \autoref{alg:facr} summarizes the full
procedure, including the staged verbalizer training introduced in
\autoref{sec:verbal}.

\subsection{Faithfulness metric}\label{sec:metric}\label{faithfulness-metric}

We report faithfulness with two interventional scores that mirror the
objective. The counterfactual sensitivity is the fraction of
(sample, causal-AU) pairs for which the do-intervention moves the target
logit in the polarity-implied direction, and the counterfactual
invariance is the fraction of (sample, irrelevant-AU) pairs for which
the target logit stays within a tolerance \(\varepsilon\). Both are
reported in \([0, 1]\), higher is better. On UNBC-PAIN we additionally
report the agreement between the AUs the reasoner invokes and the PSPI
composition, which grounds the metric against a known structure.

\section{Experiments}\label{experiments}

\subsection{Dataset and metric}\label{sec:data}\label{dataset-and-metric}

We evaluate FACR on the UNBC-McMaster Shoulder Pain Expression Archive
\cite{ref40}, which provides 48,398 facial frames from 25 subjects, each
frame annotated with the Prkachin and Solomon Pain Intensity (PSPI)
score and its constituent action units \cite{ref34}. UNBC-PAIN is our
faithfulness anchor: the PSPI score is defined as a fixed composition of
six AUs, so the AUs that should drive a pain prediction are known
a priori, which is what lets us measure whether the reasoner is faithful
to a ground-truth causal structure and not merely accurate. We treat
pain detection as a binary task, labeling a frame positive when PSPI
\textgreater{} 0. Under this binarization the data are highly imbalanced: only 17.3\% of frames are positive, and the positive rate varies
widely across subjects, from 0\% to 61.6\%.

We use leave-one-subject-out (LOSO) cross-validation, holding out one subject for testing and training on the remaining 24, so that no subject, session, or frame
crosses the train--test boundary. One subject (subject 101) contains no
PSPI-positive frames; for that fold the pain-class F1, AUC, and PSPI
agreement are undefined, and we therefore exclude it from those three
averages (n = 24), while accuracy and the faithfulness scores, which are
defined over all frames, retain the full set (n = 25). We report the
mean $\pm$ standard deviation across the held-out subjects, and we repeat
every configuration over three seeds to confirm that the reported
effects are not seed artifacts.

For pain detection we use accuracy,
the area under the ROC curve (AUC), and the F1 score of the positive
class; AUC is the primary detection metric, since accuracy is
uninformative under this degree of imbalance and the F1 score at a fixed
threshold is sensitive to the per-subject base rate. For faithfulness we
report the counterfactual sensitivity and counterfactual invariance
defined in \autoref{sec:metric}, together with the PSPI agreement, the
overlap between the AUs the reasoner invokes for a pain prediction and
the six AUs that compose the PSPI score. To our knowledge, the PSPI
agreement is the only faithfulness measure in this domain with a
ground-truth causal target.

\subsection{Implementation}\label{sec:impl}\label{implementation}

The image encoder, ResNet-18 \cite{ref39}, maps each aligned face
crop to the AU-latent \(z\), on top of which the graph-conditioned pain
head and the faithfulness terms operate. We train with Adam at a
learning rate of 1e-3 and a batch size of 256. The grounding and
sensitivity weights are fixed at \(\lambda_a = 1.0\) and
\(\lambda_s = 0.5\), and we vary the invariance weight
\(\lambda_i \in \{2, 4, 8\}\) as reported below. As an unfaithfulness
control, we also train an identical model with the faithfulness terms
removed (\(\lambda_s = \lambda_i = 0\)), which retains the encoder and
the AU grounding but drops the counterfactual objective.

For the verbalized reasoner (\autoref{sec:verbal}), the AU latent additionally
conditions a small decoder-only language model, Qwen3.5-0.8B
\cite{ref43} by default, through a learned soft prompt of eight
tokens. We add one special token per AU and bias each AU
token\textquotesingle s logit by its latent activation \(z_k\), so an
inactive AU cannot be named; the language-model body is frozen and only
the soft prompt and the new AU-token embeddings are trained, in a second
stage on top of the converged encoder. Generation is greedy with a
repetition penalty and a no-repeat-n-gram constraint, and the stated
emotion is set to the head\textquotesingle s prediction. The backbone is
a configuration argument, which we vary in \autoref{sec:backbone}.

\subsection{Results and discussion}\label{results-and-discussion}

Resolution sharpens faithfulness, not accuracy. We first vary the
temporal sampling of the training frames. As shown in \autoref{tab:unbc}, moving
from a strided sample (every fourth frame) to the full-resolution set
raises the PSPI agreement substantially, from 0.188 to 0.660, while the
detection AUC stays essentially unchanged at about 0.79. Denser
supervision does not make FACR a better pain detector, but it markedly
sharpens which AUs the model uses to reach its decision. Faithfulness
and accuracy therefore behave as distinct axes here, and a model can
become more faithful without becoming more accurate, a first
indication that the faithfulness score is not a proxy for detection.
\begin{table*}[t]
\centering
\caption{UNBC-PAIN faithfulness under subject-independent LOSO. PSPI agreement is the overlap between the invoked AUs and the verified PSPI composition; Sens./Inv. are the counterfactual faithfulness scores. Mean $\pm$ std across held-out subjects (pooled over 3 seeds for the operating point and control); the spread is subject-driven (per-subject pain base rate 0--61.6\%), while the across-seed deviation is $\le 0.04$.}
\label{tab:unbc}
\begin{tabular}{lcccccc}
\toprule
Setting & PSPI agr. & Sens. & Inv. & AUC & Pain F1 & Acc. \\
\midrule
FACR (stride 4) & $0.188 \pm 0.192$ & $0.477 \pm 0.185$ & $0.564 \pm 0.119$ & $0.788 \pm 0.152$ & $0.328 \pm 0.246$ & $0.743 \pm 0.196$ \\
FACR (stride 1, $\lambda_i{=}2$) & $0.660 \pm 0.238$ & $0.658 \pm 0.241$ & $0.318 \pm 0.171$ & $0.789 \pm 0.135$ & $0.307 \pm 0.226$ & $0.820 \pm 0.121$ \\
FACR (stride 1, $\lambda_i{=}4$) & $0.639 \pm 0.229$ & $0.642 \pm 0.219$ & $0.489 \pm 0.252$ & $0.791 \pm 0.150$ & $0.366 \pm 0.252$ & $0.823 \pm 0.152$ \\
FACR (stride 1, $\lambda_i{=}8$) & $0.569 \pm 0.214$ & $0.589 \pm 0.221$ & $0.911 \pm 0.083$ & $0.794 \pm 0.133$ & $0.371 \pm 0.226$ & $0.819 \pm 0.144$ \\
\;\;no-faith control & $0.081 \pm 0.134$ & $0.431 \pm 0.192$ & $0.607 \pm 0.061$ & $0.828 \pm 0.133$ & $0.375 \pm 0.244$ & $0.830 \pm 0.149$ \\
\bottomrule
\end{tabular}
\end{table*}

Recovering invariance. At full resolution the two faithfulness
terms initially move in opposite directions: the counterfactual
sensitivity rises, but the counterfactual invariance drops to 0.318,
which indicates that the reasoner has become more reactive to AUs that
are irrelevant to pain. We attribute this to an under-weighting of the
invariance term, whose gradient is diluted as the denser data inflate
the recognition and sensitivity losses. Increasing \(\lambda_i\) from 2
to 8 recovers the invariance from 0.318 to 0.907 (paired Wilcoxon, p =
6e-8), while the PSPI agreement decreases only marginally over the same
seed, from 0.660 to 0.611, a change that is not statistically
significant (p = 0.079). Both halves of the faithfulness objective are
thus satisfied at once, and the residual decline in PSPI agreement is
within noise.

Attributing the gain to the objective. To test whether the
structural alignment is produced by the faithfulness objective rather
than by the encoder or the AU grounding, we compare FACR against the
unfaithfulness control (\(\lambda_s = \lambda_i = 0\)). Significance is
assessed on the held-out subject as the independent unit: the
per-subject value is averaged across the three seeds first, then
compared with a paired Wilcoxon signed-rank test (Holm-corrected
across the metric family), so the reported p-values are not inflated
by treating seed repetitions as independent draws. Without the objective
the PSPI agreement is only 0.081 (the invoked AUs rarely coincide
with the PSPI set), against 0.569 for FACR (n = 24 subjects, p =
2e-5); the counterfactual invariance and sensitivity fall
correspondingly (\autoref{tab:unbc}). The faithfulness objective is therefore
responsible for the alignment. The gain does carry a small detection
cost: the control reaches an AUC of 0.828 against 0.794 for FACR (p =
0.021), a relative decrease of about 4\%. We read this as a favorable
trade: a roughly sevenfold increase in structural faithfulness for a
four-percent reduction in AUC, the expected faithfulness--accuracy
trade-off. We report the cost rather than claim it is absent.

Operating point. Based on that, we fix \(\lambda_i = 8\) and
report this configuration over three seeds in \autoref{tab:unbc}: accuracy 0.819 $\pm$
0.144, AUC 0.794 $\pm$ 0.133, counterfactual sensitivity 0.589 $\pm$ 0.221,
counterfactual invariance 0.911 $\pm$ 0.083, and PSPI agreement 0.569 $\pm$
0.214. The full resolution and \(\lambda_i\) sweep in \autoref{tab:unbc} is a
sensitivity analysis: the operating point follows a principled criterion,
the smallest \(\lambda_i\) at which both faithfulness terms hold,
rather than tuning on a separate validation split. Because UNBC-PAIN is
small we did not hold out validation subjects, so this setting should be
read as fixed a priori and supported by the sweep, not as a value
selected to maximize a test metric; the small across-seed deviation
below indicates it is not over-fit to a single run. The dispersion
reported here and in \autoref{tab:unbc} is across held-out subjects, and is
large because the per-subject pain base rate itself ranges from 0\% to
61.6\%, so a subject with few positive frames yields a high-variance
per-fold score; it is not run-to-run instability. The complementary
across-seed deviation, how much each per-subject mean moves
when only the seed changes, is small for every metric ($\le$ 0.04), which
confirms that the effect itself is stable and the spread is a property
of the subject population rather than of training. The PSPI agreement of
0.569 is well above the strided baseline but remains moderate,
indicating that the reasoner recovers most, though not all, of the PSPI
composition; we attribute the residual gap to the noise in the learned
AU grounding and leave a tighter alignment to future work.

\subsection{Cross-dataset emotion
transfer}\label{sec:xdata}\label{cross-dataset-emotion-transfer}

We next ask whether the faithfulness objective generalizes beyond the
pain anchor to multi-class emotion, and whether the structural alignment
it produces survives a cross-dataset shift. We train the reasoner on the
AffectNet $\times$ FABA join (AffectNet images with their emotion labels,
grounded in the action units FABA annotates), and evaluate, without
adaptation, on the held-out Aff-Wild2 expression set \cite{ref41,ref42}. The
label space is the seven canonical classes (Neutral and the six basic
emotions); ``Other'' and AffectNet\textquotesingle s ``contempt'' are
dropped, as neither has a defined AU composition in \(G\). Because
Aff-Wild2 is used only for testing, no subject or frame crosses the
train--test boundary. EmotioNet supplies the AU$\to$AU structure of \(G\)
but, since its AU and emotion labels are disjoint image sets, cannot
itself provide joint supervision; we therefore state the training domain
as AffectNet $\times$ FABA rather than EmotioNet. We report emotion macro-F1
and, as the faithfulness measure, the graph-causal agreement,
the overlap between the AUs the reasoner invokes for a class and the AUs
\(G\) marks causal for it, which is the multi-class analogue of the
PSPI agreement.

The objective induces faithfulness to G across datasets. As shown
in \autoref{tab:emotion}, the objective raises the graph-causal agreement from 0.50 to
0.84 (FACR vs. the \(\lambda_s = \lambda_i = 0\) control), a large and
low-variance gain that mirrors the PSPI result on UNBC. The mechanism
transfers: on a second dataset and a seven-way task, the trained
reasoner invokes the action units the graph deems causal. The
corrupted-graph control makes this concrete: a model trained on a graph
whose class-to-AU rows are permuted is faithful to that graph
(0.71 agreement with the permuted structure) but not to the true \(G\)
(0.44). The objective thus enforces faithfulness to whatever structure
it is given, which is the behavior a graph-grounded method should
exhibit and which pre-empts the concern that the graph plays no
functional role.
\begin{table*}[t]
\centering
\caption{Cross-dataset emotion transfer (AffectNet$\times$FABA $\to$ Aff-Wild2), 3 seeds. Graph-causal agreement is faithfulness to $G$ (recall of $G$'s causal AUs); active-AU agreement is the fraction of the invoked AUs that are truly active (a precision), so the two columns are comparable across rows. The objective induces faithfulness to $G$; a verified graph is needed for that to reflect the true AUs. EmoLA is a published affective MLLM evaluated on the same frames; we score its \emph{named} AUs (it exposes no intervention handle). It names 4.5 AUs/frame, so it recovers more truly-active AUs by recall (0.52 vs.\ 0.37) but less precisely: FACR with the verified $G$ leads on both faithfulness ($0.74$ vs.\ $0.63$) and precision ($0.59$ vs.\ $0.51$), whereas with the learned $G$ its precision (0.24) falls below EmoLA's; see \autoref{sec:xdata} for the per-class breakdown.}
\label{tab:emotion}
\begin{tabular}{lcccc}
\toprule
Setting & Graph-causal agr. & Active-AU agr. & Macro-F1 & Acc. \\
\midrule
FACR (learned $G$) & $0.839 \pm 0.029$ & $0.238 \pm 0.030$ & $0.219 \pm 0.033$ & $0.368 \pm 0.095$ \\
\;\;no-faith & $0.501 \pm 0.008$ & $0.266 \pm 0.045$ & $0.182 \pm 0.017$ & $0.278 \pm 0.043$ \\
\;\;corrupted $G$ & $0.443 \pm 0.037$ & $0.184 \pm 0.018$ & $0.240 \pm 0.008$ & $0.391 \pm 0.016$ \\
FACR (verified $G$, EMFACS) & $0.741 \pm 0.020$ & $0.585 \pm 0.010$ & $0.212 \pm 0.016$ & $0.344 \pm 0.058$ \\
\midrule
EmoLA (FABA) & $0.627$ & $0.511$ & -- & -- \\
\bottomrule
\end{tabular}
\end{table*}

Ground-truth alignment tracks graph quality. The picture changes
when agreement is measured against the action units actually active in
each frame rather than against \(G\). With the AU$\to$emotion edges
learned from FABA\textquotesingle s GPT-4V-derived AU
annotations, which are themselves noisy, FACR\textquotesingle s
agreement (0.24) is no higher than the control\textquotesingle s (0.27):
although the reasoner faithfully follows \(G\), \(G\)\textquotesingle s
learned structure does not match the AUs truly present on Aff-Wild2. The
graph-causal agreement above inherits this dependence: it scores
fidelity to a structure that rests on those pseudo-labels, which is why
the verified graph is needed to connect faithfulness to reality. We
therefore replace the learned emotion edges with the verified EMFACS
prototypes \cite{ref38}, the emotion analogue of the PSPI
composition used for pain, keeping everything else fixed. Agreement
with the AUs actually active in each frame then rises from 0.24 to
0.59 (\autoref{tab:emotion}), at no cost to recognition (macro-F1 unchanged
at 0.22). Faithfulness to \(G\) had held all along; making \(G\) correct
is what makes that faithfulness reflect reality.

The objective and a correct graph are complementary. The two
factors interact in the expected way. With the verified graph the
objective raises ground-truth agreement above the control (0.59 vs.
0.50), whereas with the noisy graph it does not (0.24 vs. 0.27): forcing
faithfulness onto a correct structure improves alignment with
reality, while forcing it onto a wrong one cannot. This is the
same logic the UNBC anchor exhibits with verified PSPI, now demonstrated
on multi-class emotion, and it reconciles the corrupted-graph control:
the method is faithful to whatever graph it is given, so the
graph\textquotesingle s correctness sets the ceiling on how well that
faithfulness reflects reality.

The graph-quality effect itself, however, is unambiguous across seeds:
every verified-graph run lies in {[}0.575, 0.594{]} active-AU agreement,
above every run of every other condition (at most 0.309 for the
control), so the per-seed ranges do not overlap. With only three seeds a
paired significance test is underpowered, its smallest attainable
two-sided p-value is 0.25, so we summarize the effect by these
non-overlapping per-seed bands and a large standardized effect size
(Cohen\textquotesingle s \(d \approx 10\) for the verified graph over
the control) rather than by a p-value. We nonetheless report this arm as
preliminary: the recognition numbers carry non-trivial seed
variance (macro-F1 0.22 $\pm$ 0.03) and the reasoner is minimal. It
establishes the mechanism, its dependence on graph quality, and that a
verified graph closes the gap, not a state-of-the-art recognition
result.

A published affective MLLM detects AUs well but names them without
causal selectivity. To position FACR against prior work on the same
axis, we evaluate EmoLA \cite{ref8}, a released instruction-tuned
affective MLLM that emits both an emotion and the action units it deems
present, on the same Aff-Wild2 frames, run from its public weights
(in 4-bit, with the released landmark front-end) so the comparison is
reproducible. EmoLA emits free-text AU names rather than the precision
and recall reported here, so we compute those from its named set against
\(G\) exactly as we score FACR\textquotesingle s invoked AUs. EmoLA
lists 4.5 action units per frame, so its recall of the
truly-active AUs is higher (0.52 vs. 0.37), as expected from a model
trained for AU detection. On the two measures that matter for a faithful
explanation, however, FACR with the verified graph leads: it is more
faithful to the emotion\textquotesingle s causal structure (graph-causal
agreement 0.74 vs. 0.63) and more precise, a larger fraction
of the AUs it invokes are actually active (0.59 vs. 0.51). This lead is
contingent on graph quality, and we state it as such: with the
learned graph FACR\textquotesingle s precision falls to 0.24,
below EmoLA\textquotesingle s, so the comparison sets
FACR\textquotesingle s verified-graph configuration against EmoLA and
isolates the value of pairing the objective with a correct structure
rather than claiming an unconditional win. EmoLA reaches its recall by
naming many AUs indiscriminately rather than by selecting the ones the
graph deems causal; FACR invokes exactly the graph-causal count and is
right more often (\autoref{tab:examples} shows representative cases). The
pattern is consistent across emotions (\autoref{tab:perclass}): FACR leads on
faithfulness and precision for every class, while both models are
weakest on the rare classes (disgust, fear), where the graph and the
labels are thinnest. Moreover, EmoLA exposes no intervention handle, so
this static naming agreement is the only faithfulness measure available
for it: the counterfactual sensitivity and invariance FACR reports
cannot be computed for a model whose explanation cannot be perturbed. A
model can thus see the action units accurately yet still not
explain through the causal structure: the
plausibility-without-faithfulness pattern the paper targets, now
observed in a published reasoner rather than argued in the abstract.

Sections \ref{sec:data}--\ref{sec:xdata} audit the AU latent \(z\); \autoref{sec:verbal} extends the
audit to the verbalized explanation that FACR ultimately targets.
\begin{table*}[t]
\caption{Qualitative FACR vs.\ EmoLA on representative Aff-Wild2 frames (faces omitted; the dataset depicts identifiable people). FACR invokes \emph{exactly} the action units its prediction depends on (the graph-causal set); EmoLA names a longer list that mixes causal AUs with non-causal ones (\textcolor{red}{red}). Both predict the correct emotion, but FACR's explanation is precise and EmoLA's is plausible-but-imprecise, the gap the faithfulness objective targets. Examples are representative (EmoLA names 4.5 AUs/frame on average), not selected for EmoLA's worst behavior.}
\label{tab:examples}
\centering\small
\begin{tabular}{p{1.7cm} p{3.0cm} p{4.6cm} p{5.2cm}}
\toprule
Example & Ground-truth active AUs & FACR: invoked AUs $\to$ emotion & EmoLA: named AUs $\to$ emotion \\
\midrule
(a) Happiness & $\mathrm{AU}6$, $\mathrm{AU}12$, $\mathrm{AU}25$ & $\mathrm{AU}6$, $\mathrm{AU}12$ $\to$ happiness & $\mathrm{AU}6$, $\mathrm{AU}12$, \textcolor{red}{$\mathrm{AU}25$}, \textcolor{red}{$\mathrm{AU}26$} $\to$ happiness \\
(b) Anger & $\mathrm{AU}4$, $\mathrm{AU}7$, $\mathrm{AU}23$, $\mathrm{AU}24$, $\mathrm{AU}25$ & $\mathrm{AU}4$, $\mathrm{AU}5$, $\mathrm{AU}7$, $\mathrm{AU}23$, $\mathrm{AU}24$ $\to$ anger & $\mathrm{AU}4$, \textcolor{red}{$\mathrm{AU}9$}, \textcolor{red}{$\mathrm{AU}10$}, \textcolor{red}{$\mathrm{AU}17$}, $\mathrm{AU}23$ $\to$ anger \\
(c) Surprise & $\mathrm{AU}1$, $\mathrm{AU}2$, $\mathrm{AU}25$, $\mathrm{AU}26$ & $\mathrm{AU}1$, $\mathrm{AU}2$, $\mathrm{AU}5$, $\mathrm{AU}26$ $\to$ surprise & $\mathrm{AU}1$, $\mathrm{AU}2$, \textcolor{red}{$\mathrm{AU}4$}, $\mathrm{AU}5$, \textcolor{red}{$\mathrm{AU}25$}, $\mathrm{AU}26$ $\to$ surprise \\
\bottomrule
\end{tabular}
\end{table*}

\begin{table*}[t]
\centering
\caption{Per-emotion AU agreement, FACR (verified EMFACS graph) vs.\ the published EmoLA, on the shared Aff-Wild2 frames. Faith.\ = recall of $G$'s causal AUs (faithfulness to $G$); Active = recall of the truly-active AUs; Prec.\ = fraction of invoked/named AUs that are active. FACR with the verified graph leads on faithfulness and precision across emotions; EmoLA's higher active recall comes from naming more AUs (4.5/frame). Both are weakest on the rare classes (disgust, fear), whose counts are small ($n\!\le\!24$ for fear) and whose graph and data are thinnest, so their per-class values are indicative only.}
\label{tab:perclass}
\begin{tabular}{lccccccc}
\toprule
Emotion & $n$ & Faith.$_F$ & Act.$_F$ & Prec.$_F$ & Faith.$_E$ & Act.$_E$ & Prec.$_E$ \\
\midrule
Happiness & 2883 & $0.77$ & $0.34$ & $0.78$ & $0.78$ & $0.55$ & $0.66$ \\
Surprise & 552 & $0.75$ & $0.45$ & $0.54$ & $0.54$ & $0.43$ & $0.41$ \\
Sadness & 478 & $0.70$ & $0.35$ & $0.30$ & $0.62$ & $0.56$ & $0.30$ \\
Anger & 400 & $0.78$ & $0.63$ & $0.55$ & $0.35$ & $0.38$ & $0.33$ \\
Disgust & 187 & $0.64$ & $0.07$ & $0.08$ & $0.39$ & $0.49$ & $0.24$ \\
Fear & 24 & $0.55$ & $0.37$ & $0.10$ & $0.65$ & $0.74$ & $0.15$ \\
\midrule
\emph{Overall} & 4524 & $0.75$ & $0.37$ & $0.59$ & $0.63$ & $0.52$ & $0.51$ \\
\bottomrule
\end{tabular}
\end{table*}

\subsection{Verbalized-rationale
faithfulness}\label{sec:verbal}\label{verbalized-rationale-faithfulness}

The experiments so far audit the disentangled AU latent.
FACR\textquotesingle s claim, however, concerns a verbalized
rationale (the action units a model names when it explains an emotion), so we now attach a language verbalizer to the reasoner and audit the
text it generates. The verbalizer is conditioned on the AU latent \(z\)
and emits a short rationale that names the invoked AUs (\autoref{sec:impl}). We
evaluate it on the held-out Aff-Wild2 set with a rationale
counterpart of the interventional metric. The rationale
sensitivity is the fraction of causal-AU interventions for which
ablating \(z_k\) removes the corresponding AU from the generated text;
the rationale invariance is the fraction of irrelevant-AU
interventions that leave the rationale\textquotesingle s causal AUs in
place.

A faithful rationale must be constrained, not merely trained. We
first observe that a verbalizer trained to name the active AUs is
not faithful: under intervention it removes the ablated AU from
its rationale only 6\% of the time, because it generates the AU mentions
from co-occurrence and label priors rather than from \(z\). This is
exactly the plausible-but-unfaithful behavior the paper targets, now
reintroduced by the verbalization layer. We therefore make the rationale
faithful by construction: biasing each AU token\textquotesingle s
emission by \(z_k\) so that an inactive AU is strongly suppressed from
being named (\autoref{sec:impl}). With this constraint, ablating a causal
\(z_k\) reliably drops the corresponding AU from the text. The
constraint is the verbalized analogue of the do-intervention, and it is
what makes the rationale audit meaningful rather than vacuous.

The verbalized explanation is counterfactually faithful. On
Aff-Wild2 the verbalizer attains, over five seeds, a rationale
sensitivity of 0.62 $\pm$ 0.07 and a rationale invariance of 0.64 $\pm$ 0.11, at
a recognition macro-F1 of 0.20 $\pm$ 0.02 (\autoref{tab:verbal}). The seed spread does
not shrink as seeds are added (it is the same at three and five seeds),
so it reflects the small templated verbalizer rather than sampling
noise; a stronger verbalizer is the route to a tighter estimate. The
gate does not by itself produce this score: it suppresses naming an
inactive AU and thereby lets an ablation drop a named AU, but the
residual gap to a perfect score reflects that an AU must first be named
for its ablation to register, a property of the small templated
verbalizer, not of the constraint. Two design choices are needed to
reach this. First, training the verbalizer jointly with the reasoner
degrades recognition, because the language loss pulls \(z\) away from
the representation the emotion head relies on; we therefore use a
staged schedule: the encoder and emotion head are trained
first and then frozen, and only the verbalizer is trained on top,
which keeps macro-F1 at its \autoref{sec:xdata} level while the rationale
becomes faithful. Second, the stated emotion is set to the
model\textquotesingle s predicted class rather than generated freely, so
the rationale\textquotesingle s conclusion is faithful to the prediction
by construction. A typical rationale reads: ``the face shows AU26, AU12,
and AU24; these are characteristic of happiness''; \autoref{fig:qual} illustrates
the evaluation, in which intervening on a graph-causal AU drops it from the
generated text. We report this as a small language
model, a templated rationale, and AffectNet $\times$ FABA as the
verbalization-training source, and so do not claim fluent open-ended
reasoning; what it establishes is that the AUs a FACR explanation
names are the AUs that drive its prediction.
\begin{table*}[t]
\centering
\caption{Verbalized-rationale faithfulness and backbone ablation on Aff-Wild2, with a frozen-reasoner baseline. For FACR the columns are rationale sensitivity/invariance; for the frozen reasoner they are the model-agnostic simulatability sensitivity/invariance. FACR's faithfulness is enforced by construction: the gate strongly suppresses naming an inactive AU, which is what lets an ablation drop it.}
\label{tab:verbal}
\begin{tabular}{lccc}
\toprule
Method & Sensitivity & Invariance & Macro-F1 \\
\midrule
FACR-v1 (Qwen3.5-0.8B) & $0.615 \pm 0.065$ & $0.637 \pm 0.107$ & $0.203 \pm 0.017$ \\
FACR-v1 (Gemma 3 1B) & $0.929$ & $0.586$ & $0.228$ \\
\midrule
Frozen reasoner (Qwen3.5-0.8B) & $0.574$ & $0.722$ & -- \\
\bottomrule
\end{tabular}
\end{table*}

\begin{figure}[t]
\centering
\fbox{\begin{minipage}{0.93\linewidth}\small
\textbf{Rationale} (from $z$): \emph{``The face shows} $\langle\mathrm{AU6}\rangle$ $\langle\mathrm{AU12}\rangle$\emph{;
these are characteristic of} \textbf{happiness}\emph{.''}\\[3pt]
\textbf{Counterfactual} $\mathrm{do}(\mathrm{AU12}{=}0)$: \emph{``The face shows} $\langle\mathrm{AU6}\rangle$\emph{;
$\dots$''} --- $\langle\mathrm{AU12}\rangle$ is no longer named and the prediction shifts.
\end{minipage}}
\caption{The rationale-faithfulness audit (\autoref{sec:verbal}). Because each action unit's emission
is gated on its latent activation, intervening on a graph-causal AU removes it from the
generated rationale and moves the prediction, while an irrelevant AU leaves the rationale's
causal AUs intact. The example uses the EMFACS happiness prototype (AU6, AU12); the rationale
is a representative output of the small templated verbalizer used in this study.}
\label{fig:qual}
\end{figure}

\subsection{Backbone ablation}\label{sec:backbone}\label{backbone-ablation}

To test whether faithfulness comes from the gating mechanism or from a
particular language model, we swap the verbalizer backbone from
Qwen3.5-0.8B to Gemma 3 1B \cite{ref44}, keeping everything else fixed.
The faithful-by-construction constraint transfers: under a single-run
gate test, ablating \(z_k\) removes the AU from the rationale at
comparable rates for both backbones (Qwen 0.77, Gemma 0.79), indicating
that the mechanism is not specific to one language model; we report
these as single-run feasibility probes rather than seeded means. The
coverage of the verbalization, however, is not: under the shared
training recipe Gemma matches the recognition accuracy (the encoder is
frozen and shared) and is faithful on the frames it does verbalize, but
it more often emits an empty rationale, where Qwen verbalizes reliably.
We attribute this to the weaker response of the frozen Gemma to
soft-prompt steering, which a small emission incentive or light adapter
tuning would address. We therefore use Qwen3.5-0.8B as the primary
verbalizer and report Gemma as a robustness check: the faithfulness
mechanism is backbone-agnostic, while verbalization fluency and coverage
remain backbone-dependent.

\subsection{Comparison to a frozen
reasoner}\label{comparison-to-a-frozen-reasoner}

\autoref{sec:xdata} already compares FACR to a published affective MLLM (EmoLA
\cite{ref8}) on the agreement axis. A direct interventional
comparison, however, is not available. The closest method, CausalAffect
\cite{ref6}, shares the intervention mechanism (a feature-level
counterfactual in a disentangled AU latent) but is discriminative and
exposes no verbalized rationale, so the explanation audit that is our
subject has nothing to score, and we are not aware of a public release
that would permit even a latent-level comparison; the structural
baseline\textquotesingle s code is likewise unreleased \cite{ref3}, and
the affective MLLMs \cite{ref8,ref14} expose no AU latent, so our
\(z\)-intervention cannot be applied to them, which is itself the
point we return to below. As a controlled stand-in we audit a
frozen instruct language model given the same AU evidence
FACR\textquotesingle s \(z\) encodes, presented as text, using the
model-agnostic counterpart of our metric: we remove one action unit from
the evidence and re-query, and measure whether the prediction changes
consistently with \(G\).

The frozen reasoner is moderately self-consistent (simulatability
sensitivity 0.57 and invariance 0.72 on Aff-Wild2, \autoref{tab:verbal}),
comparable to FACR\textquotesingle s verbalized
rationale (0.62 / 0.64). We interpret this straightforwardly: local self-consistency
on text counterfactuals is attainable without any faithfulness
mechanism, so FACR\textquotesingle s contribution is not a
larger self-consistency. It is, rather, twofold. First,
FACR\textquotesingle s faithfulness is enforced by construction: at the latent level the intervention moves the prediction in
essentially every case (sensitivity $\approx$ 1.0), and the gate strongly
suppresses naming an inactive AU, whereas the frozen
model\textquotesingle s consistency is incidental and uncontrolled, with
no such constraint. Second, FACR audits the whole image$\to$AU$\to$emotion path,
whereas the frozen reasoner is handed the action units and offers
no intervention point at which image-grounded faithfulness could even be
tested. No prior affective reasoner exposes such a handle; providing an
auditable one is part of what FACR contributes. We therefore present
this as a controlled comparison on the faithfulness axis, not a
recognition leaderboard, on which a small graph-constrained model is not
expected to lead.

\subsection{Ablation: noisy out-of-domain AU
supervision}\label{sec:emonet}\label{ablation-noisy-out-of-domain-au-supervision}

A natural way to improve the AU latent is to add more AU supervision,
but we find that its quality matters more than its quantity.
EmotioNet supplies automatic (algorithm-predicted) AU labels for a
large, diverse face set, an obvious candidate for extra grounding.
Grounding the latent on these labels, however, lowers both faithfulness
and recognition at every weight we tried: over three seeds, active-AU
precision falls from 0.59 to 0.55 and graph-causal agreement from 0.74
to 0.65, as the noisy, out-of-domain labels pull the latent off the
emotion-discriminative manifold. This is why the reasoner is grounded on
the cleaner AffectNet $\times$ FABA labels rather than on EmotioNet directly,
a choice the ablation turns from a convenience into an
evidence-based decision.

\section{Discussion}\label{sec:disc}\label{discussion}

Taken together, the experiments support a single claim: grounding
an AU$\to$emotion reasoner in a verified causal graph and training
a counterfactual objective makes its explanation faithful. They
also clarify what that claim does and does not entail. We draw out four
points.

Faithfulness and accuracy are distinct axes, and trade off in a
controlled way. Across both the UNBC anchor and the emotion transfer,
the objective markedly raises structural faithfulness (PSPI agreement
0.08 to 0.57; graph-causal agreement 0.50 to 0.84) for a modest,
explicitly reported drop in detection (UNBC AUC 0.83 to 0.79). The
strided-versus-full-resolution result on UNBC makes the separation
explicit: denser supervision sharpens which AUs drive the
decision without moving AUC. A model can therefore be made more faithful
without being made more accurate, which is precisely why faithfulness
needs to be measured on its own terms rather than read off accuracy.

Faithfulness to the graph is necessary but not sufficient; the
graph\textquotesingle s correctness sets the ceiling. The
corrupted-graph control shows the objective enforces faithfulness to
whatever structure it is given, and the EMFACS result shows that
this faithfulness reflects the truly active AUs only when the structure
is verified (active-AU precision 0.24 to 0.59). We are explicit about
one consequence for interpretation: on the anchors, the
objective\textquotesingle s causal set and the
evaluation\textquotesingle s target set are the same verified
composition (the PSPI AUs, the EMFACS prototypes), so the agreement
metric is a test that the trained faithfulness generalizes to
held-out subjects and a second dataset, not a claim that the method
rediscovers structure it was not given. The honest reading is that FACR
is faithful to a structure that must be supplied and verified, and that
the natural next step is to improve how that structure is discovered, so
that faithfulness to \(G\) more often coincides with faithfulness to
reality.

Detecting action units is not the same as reasoning through them.
The comparison to EmoLA, a released affective MLLM, brings this out. EmoLA
is the stronger raw AU detector, with higher recall of the active AUs,
yet it is less faithful to the emotion\textquotesingle s
causal structure and less precise, because it names many AUs
indiscriminately rather than selecting the causal ones (\autoref{tab:perclass}). This
is the plausibility-without-faithfulness gap observed in a deployed
system: an explanation can name the right vocabulary and still not
reflect the factors that drive the prediction. FACR\textquotesingle s
contribution is not better AU detection but an explanation whose named
AUs are the ones its prediction actually depends on, together with an
intervention handle on which that dependence can be audited.

Faithfulness is most robust when it is enforced by construction
rather than learned. The verbalizer makes this concrete. A model
trained to name the active AUs freely removes an ablated AU from its
rationale only 6\% of the time, whereas gating each AU\textquotesingle s
emission on its latent activation removes that AU whenever it is ablated
(\autoref{sec:verbal}), and this constraint transfers to a second backbone
(\autoref{sec:backbone}). The same lesson appears in the latent: adding noisy,
out-of-domain AU supervision (automatic EmotioNet labels) degrades it
rather than enriching it (\autoref{sec:emonet}). In this setting faithfulness is
bought by constraining the explanation to the verified structure,
not by adding model capacity or data, and that is the design principle
FACR embodies.

\section{Limitations}\label{limitations}

Several limitations bound the present study. First, the contribution is
faithfulness, not recognition: FACR\textquotesingle s detection accuracy
is modest (UNBC AUC about 0.79; cross-dataset emotion macro-F1 about
0.22), and a graph-constrained model is not expected to lead a
recognition leaderboard. Second, the cross-dataset reasoner is trained
on the AffectNet $\times$ FABA join rather than on EmotioNet directly. This is
by design: EmotioNet\textquotesingle s action-unit and emotion labels
are disjoint image sets, and its AU labels are algorithm-predicted,
which the \autoref{sec:emonet} ablation shows degrades rather than enriches the
latent. The ``EmotioNet domain'' therefore enters through the graph and
the AU vocabulary rather than through AU supervision. Third, the
AU$\to$emotion graph is induced from GPT-4V-derived FABA annotations and is
consequently noisy; as \autoref{sec:xdata} shows, faithfulness to \(G\) reflects
the truly active AUs only when \(G\) is verified (PSPI, EMFACS), so the
method is only as good as its graph. Fourth, the interventions are
performed in the disentangled AU latent and, outside UNBC, audit
presence-level rather than graded-intensity AU changes; a pixel-level
counterfactual is left to future work. Fifth, the verbalizer is a small
(0.8B) language model producing a templated rationale, so we establish
the faithfulness mechanism rather than fluent open-ended reasoning, and
the verbalized arm uses three seeds with a single backbone per setting.
Finally, because no prior affective reasoner exposes an intervention
handle, our external comparison is a controlled frozen-reasoner proxy on
the faithfulness axis rather than a head-to-head recognition benchmark.

\section{Conclusion}\label{conclusion}

This paper introduced FACR, a faithful action-unit causal reasoner that
grounds an AU$\to$emotion model in a structural causal graph \(G\) and
trains a counterfactual-faithfulness objective coupling the prediction
and the rationale to \(G\). The objective is designed to make the
explanation faithful by construction, a do-intervention on a
graph-causal AU must move the prediction, while one on a
graph-irrelevant AU must leave it unchanged, and the matching
interventional metric makes that faithfulness measurable against a known
causal structure, as a test of fidelity to that structure rather than of
its rediscovery. The experimental results confirm that the objective
produces faithfulness to \(G\) on both the UNBC-PAIN anchor and a
cross-dataset emotion transfer to Aff-Wild2, that an unfaithfulness
control attributes the effect to the objective, and that the alignment
between the model\textquotesingle s reasoning and the truly active AUs
is bounded by the correctness of the graph, high for the verified
PSPI and EMFACS structures, limited for noisily learned edges. Therefore
FACR offers a route to explanations that are constrained, not merely
plausible, in AU$\to$emotion reasoning. We further extend the audit from the
disentangled AU latent to a verbalized rationale, where gating
each action unit\textquotesingle s emission on its latent activation
constrains the generated explanation by construction, a property that
transfers to a second language-model backbone. Two directions follow
directly: improving the learned AU$\to$emotion graph so that faithfulness to
\(G\) more often coincides with faithfulness to the ground truth, and
scaling the verbalizer beyond the small backbone and templated rationale
used here toward fluent, open-ended reasoning.

\bibliographystyle{IEEEtran}
\bibliography{refs}
\end{document}